\documentclass[lettersize,journal]{IEEEtran}
\usepackage{amsmath,amsfonts}
\usepackage{algorithmic}
\usepackage{algorithm}
\usepackage{array}
\usepackage[caption=false,font=normalsize,labelfont=sf,textfont=sf]{subfig}
\usepackage{textcomp}
\usepackage{stfloats}
\usepackage{url}
\usepackage{verbatim}
\usepackage{graphicx}
\usepackage{cite}
\usepackage{amssymb}
\usepackage{mathtools}
\usepackage{amsthm}
\usepackage{booktabs}
\usepackage{colortbl}
\usepackage{xcolor}
\definecolor{dropcolor}{RGB}{110, 135, 160}
\usepackage{tcolorbox}
\usepackage{tabularx}
\usepackage{multirow}
\definecolor{tabfirst}{rgb}{0.7, 0.85, 0.95}
\definecolor{tabsecond}{rgb}{0.88, 0.95, 1.0}
\definecolor{tabthird}{rgb}{1, 1, 0.7}
\usepackage{microtype}
\usepackage{graphicx}
\usepackage{booktabs} 

\newcommand{\cfirst}{\cellcolor{tabfirst}\bfseries}
\newcommand{\csecond}{\cellcolor{tabsecond}}

\begin{document}

\title{VLANeXt Family: A Systematic Study of VLA Models from Core Recipes to Emerging Paradigms}

\author{
Xiao-Ming Wu,
Kang Liao,
Yihang Luo,
Bin Fan,
Jian-Jian Jiang,
Runze Yang,
Zhonghua Wu,
Wei-Shi Zheng,
and Chen Change Loy
\thanks{
Xiao-Ming Wu, Kang Liao, Yihang Luo, and Chen Change Loy are with
S-Lab, Nanyang Technological University, Singapore.
Chen Change Loy is also with ACE Robotics.
}
\thanks{
Bin Fan, Jian-Jian Jiang, and Wei-Shi Zheng are with
Sun Yat-sen University, China.
}
\thanks{
Runze Yang is with Shanghai Jiao Tong University, China.
}
\thanks{
Zhonghua Wu is with SenseTime Research, Singapore.
}
\thanks{
Corresponding author: Chen Change Loy
(e-mail: ccloy@ntu.edu.sg).
}
}



\maketitle

\begin{abstract}
Following the rise of large foundation models, Vision–Language–Action models (VLAs) emerged, leveraging strong visual and language understanding from Vision-Language Models for general-purpose policy learning. Yet, the current VLA landscape remains fragmented and exploratory. Although many groups have proposed their own VLA models, inconsistencies in training protocols and evaluation settings make it difficult to identify which design choices truly matter.
To bring structure to this evolving space, we reexamine the VLA design space under a unified framework and evaluation setup. Starting from a simple VLA baseline similar to RT-2, which is the origin of VLA, we systematically dissect design choices along three dimensions: foundational components, perception essentials, and action modeling perspectives. From this study, we distill 12 key findings that together form a \textit{practical recipe} for building strong VLA models. The outcome of this exploration is a simple yet effective model, VLANeXt. It outperforms the state-of-the-art methods on the LIBERO and LIBERO-plus benchmarks and demonstrates strong performance in real-world experiments.
Beyond identifying the core recipe, we further ask how far these design principles extend to the \textit{emerging paradigms} in VLAs. We thus expand VLANeXt along several emerging directions, including model scaling, latent-action pretraining, latent predictive representation learning, and world action modeling. These studies give rise to the VLANeXt family, spanning compact and scaled VLA variants, latent-action models, JEPA-style predictive models, and World Action Models. Our results show that the core recipe provides a strong foundation across different model scales and emerging paradigms.
We release a unified and easy-to-use codebase to reproduce our findings, explore the design space, and develop new VLA variants on top of a shared foundation. The codebase is available at \url{https://github.com/DravenALG/VLANeXt}.
\end{abstract}


\section{Introduction}

\begin{figure}[t]
  \begin{center}
    \centerline{\includegraphics[width=\columnwidth]{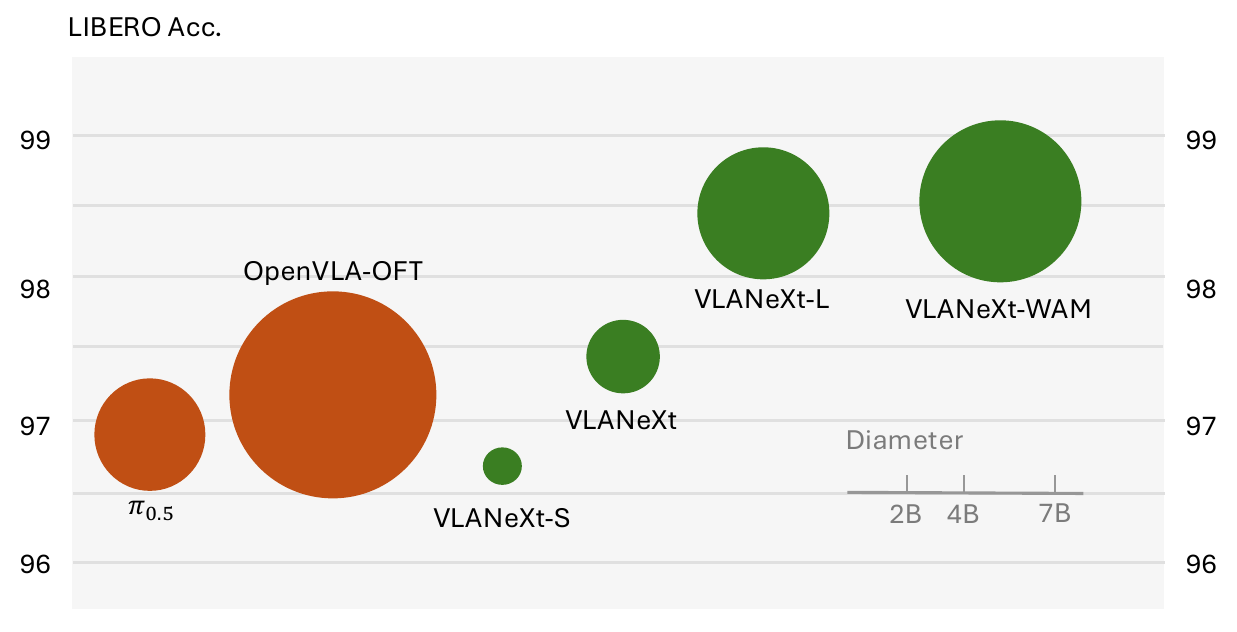}}
    \caption{
      \textbf{Performance comparison on the LIBERO benchmark}. We compare VLANeXt with representative VLA baselines across model scales. Despite its smaller model size, VLANeXt achieves higher success rates than prior methods on the LIBERO benchmark, demonstrating the effectiveness of the design recipe distilled in this work. We further extend VLANeXt to emerging paradigms through model scaling and world action modeling. Both VLANeXt-L and VLANeXt-WAM further improve upon the base model, showing that the core recipe remains effective as VLAs scale and evolve toward emerging paradigms.
    }
    \label{fig:performance_first_glance}
  \end{center}
  \vspace{-10mm}
\end{figure}

\begin{figure}[t!]
  \begin{center}
    \centerline{\includegraphics[width=0.98\columnwidth]{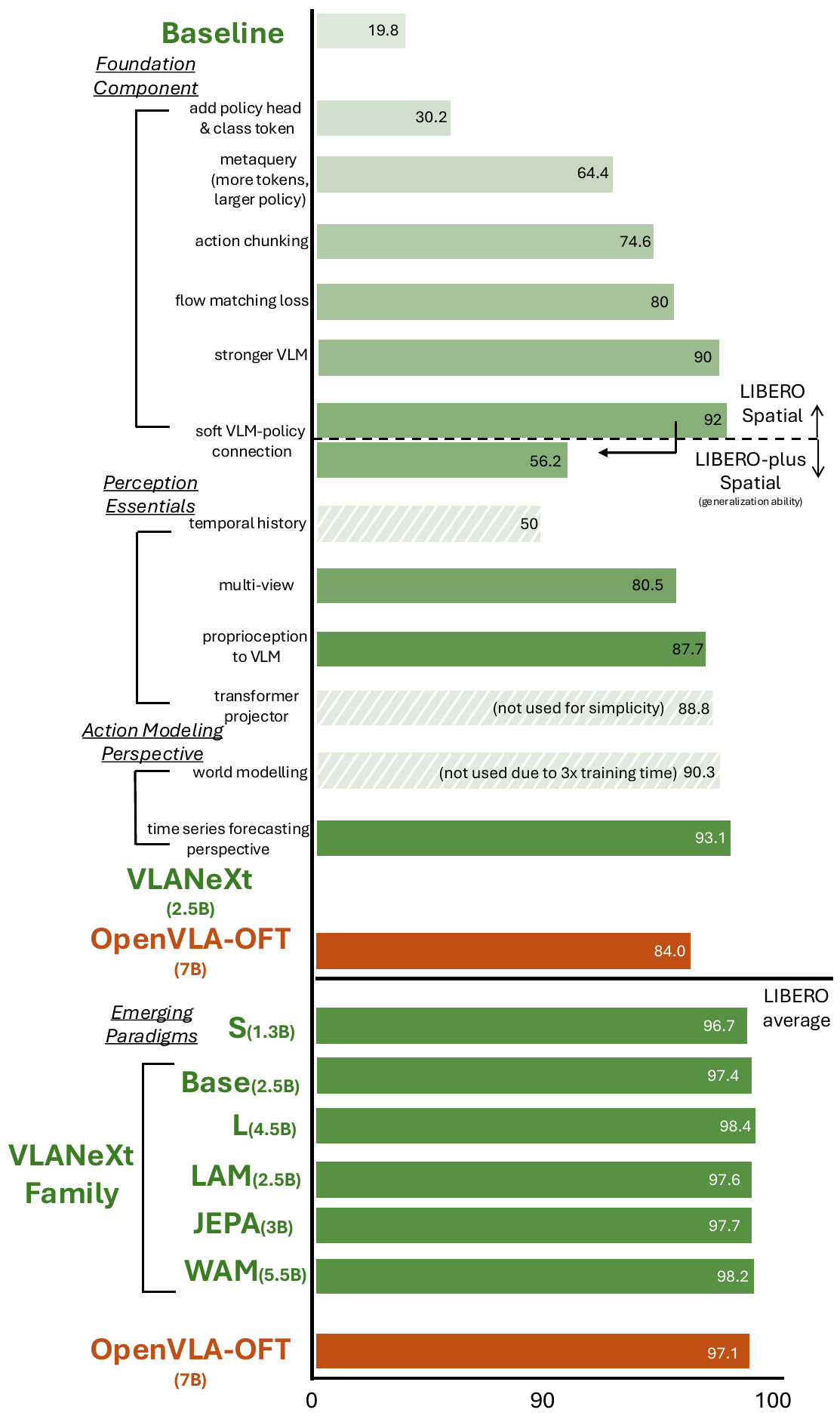}}
    \caption{
        \textbf{Ablation trajectory across the VLA design space}. We progressively evolve a baseline VLA across foundational components, perception essentials, and action modeling perspectives, evaluating on LIBERO (Spatial) and switching to LIBERO-plus (Spatial) once performance saturates. This trajectory yields the base VLANeXt (2.5B), outperforming OpenVLA-OFT (7B). We further extend VLANeXt to emerging paradigms, including model scaling (S/Base/L), latent-action pretraining (LAM), latent predictive representation learning (JEPA), and world action modeling (WAM). The resulting VLANeXt family consistently achieves strong average performance across the four LIBERO suites.
    }
    \label{fig:roadmap}
  \end{center}
  \vspace{-8mm}
\end{figure}

\IEEEPARstart{B}{oth} academia and industry have proposed a wide range of vision-language-action (VLA) models that demonstrate strong performance and encouraging generalization across diverse tasks~\cite{2023_rt2, 2024_open-x-embodiedment, 2023_flamingo, 2024_cogcot, 2024_openvla, 2024_pi0, 2025_gemini-robotcs, 2025_nora, 2025_openvla-oft, 2025_smolvla, 2025_pi0.5, 2025_pi0.6, 2026_robovlms}. Most VLA approaches follow a similar paradigm: they build on pre-trained LLMs or VLMs, processing visual observations together with language instructions to derive action-relevant representations for policy learning. This pipeline introduces numerous design choices, including how to interface the VLM with the policy module, how to train the policy, how to select essential perceptual inputs, and how actions should be represented and modeled. Despite rapid progress, early exploration of VLAs remains something of a ``primordial soup'',  rich in ideas but lacking clear structure. While prior work has explored VLA design from certain perspectives~\cite{2024_3D-VLA, 2025_spatialvla, 2025_dreamvla, 2025_worldvla, 2025_up-vla, 2025_GRAPE, 2025_vla-rl}, differences in training protocols and evaluation setups make it difficult to identify which design choices in the shared VLA design space truly matter.

This work aims to provide a more systematic understanding of this fragmented design space by comprehensively reexamining VLA designs under a unified framework and evaluation protocol. While several prior works~\cite{2025_openvla-oft, 2026_robovlms} have made preliminary attempts to explore VLA designs, their investigations remain limited in scope. In detail, we begin with a simple baseline VLA, similar to RT-2~\cite{2023_rt2}, which is the origin of VLA and serves as a strong reference point for analyzing the effectiveness of different design choices. We evaluate the core-recipe variants on two commonly used VLA benchmarks, including LIBERO~\cite{2023_libero} and LIBERO-plus~\cite{2025_libero-plus}, where LIBERO-plus extends the original benchmark with controlled and unseen perturbations to better assess robustness and generalization. After distilling the core design principles under this unified setup, we further examine how far these principles extend to emerging VLA paradigms beyond the standard formulation.

Within this setup, we first systematically explore the design space along three dimensions: 1) foundational components, covering core VLM-policy architectures and action learning objectives; 2) perception essentials, examining the role of visual, language, and proprioceptive inputs; and 3) action modeling perspectives, investigating designs and auxiliary objectives that facilitate action generation. We conduct more than 500 distinct experiments over the above three dimensions and distill 12 key findings that together form a practical recipe for building strong VLA models, summarized in Fig.~\ref{fig:roadmap}.

We highlight several findings that we believe are novel and noteworthy for the field: 1) a soft connection between the VLM and the policy module performs slightly better than both loose and tight coupling strategies; 2) simply providing temporal observations as additional VLM inputs does not improve action learning, despite the backbone being pretrained for video understanding; 3) conditioning proprioceptive input in the VLM yields better performance than either omitting proprioception or injecting it directly into the policy module; and 4) framing action generation as a time-series forecasting problem and incorporating frequency-domain modeling provides an effective and efficient way to improve action prediction.

The outcome of this study is a simple yet effective VLA model, \textbf{VLANeXt}, derived directly from the design principles uncovered in our systematic exploration. Rather than relying on aggressive model scaling or task-specific engineering, VLANeXt achieves state-of-the-art performance on both LIBERO~\cite{2023_libero} and LIBERO-plus~\cite{2025_libero-plus}, and adapts effectively to real-world manipulation tasks. These results show that strong VLA performance can emerge from principled design choices within a unified framework.

The above findings are reported in our earlier conference version~\cite{2026_vlanext}. In this extended work, we further explore how far these design principles extend to the \textit{emerging paradigms} in VLAs. We expand VLANeXt along four directions. First, we study \textit{model scaling} across different backbone sizes, examining whether the recipe remains effective as model capacity changes. Second, we investigate \textit{latent-action pretraining}, where the model first learns from latent action representations before adapting to labeled robot actions. Third, we explore \textit{latent predictive representation learning} through JEPA-style objectives, encouraging the model to predict future latent representations rather than directly reconstructing future observations. Fourth, we extend VLANeXt toward \textit{world action modeling}, jointly learning future visual dynamics and robot actions with video-generation backbones. Together, these extensions form the \textbf{VLANeXt family}, spanning compact and scaled VLA variants, latent-action models, JEPA-style predictive models, and World Action Models. Our results show that the core recipe provides a strong foundation across different model scales and emerging paradigms.

To support further progress in this direction, we release a \textbf{research-oriented codebase} that standardizes training and evaluation while exposing the key components of the VLA design space. The framework is intentionally lightweight and minimally encapsulated, enabling researchers to reproduce our findings, compare new designs under a shared protocol, and develop new variants across the broader VLANeXt family and emerging VLA paradigms. The codebase is available at \url{https://github.com/DravenALG/VLANeXt}.

\section{Revisiting Robot Learning and VLA Models}

Robot learning aims to apply machine learning techniques to robotic control, empowering robots to interact with the physical world and acquire diverse skills~\cite{2020_robot-learning}. In the context of robot learning, tasks are generally categorized into locomotion and manipulation based on the active components being controlled. Locomotion is designed to maintain the stability and balance of the robot base, enabling mobility for legged systems such as quadrupeds or humanoids~\cite{2018_deepmimic, 2021_rma, 2020_challenging-terrain, 2023_walk-these-ways, 2024_not-only-rewards, 2025_locoformer}. Since these tasks typically possess explicit objectives, they frequently use Reinforcement Learning (RL). In contrast, robotic manipulation focuses on controlling the robotic arm or whole body to execute a diverse array of interactive tasks. Previously, manipulation was usually focused on specific tasks, such as gripper grasping~\cite{2020_graspnet, 2021_graspness, 2024_economicgrasp, 2024_r2sgrasp, 2025_task-oriented}, dexterous hand grasping~\cite{2023_contactgen, 2023_contact2grasp, 2023_scenediffuser, 2024_DGTR, 2024_Grasp-as-you-say, 2025_afforddexgrasp}, non-prehensile manipulation~\cite{1999_non-prehensive-progress, 2012_contact-invariant, 2025_dywa}, pick-and-place/rearrangement~\cite{2020_form2fit, 2021_transporter}, contact-rich insertion~\cite{2015_contact-rich}, dexterous manipulation~\cite{2017_dexerous_manipulation, 2020_dexerous_in_hand_manipulation}, deformable-object manipulation~\cite{2018_deformable_manipulation}, articulated-object manipulation~\cite{2021_where2act}, dynamic manipulation~\cite{2020_tossingbot, 2023_target-reference, 2024_motiongrasp}, and tool use~\cite{2005_tool_affordance, 2017_dexerous_manipulation}. These tasks can typically be formulated with well-defined task priors, making them relatively easier to learn and generalize. More recently, general-purpose manipulation has gradually become a central research direction~\cite{2023_rt1, 2025_diffusion-policy, 2024_hpt, 2024_RDT-1B, 2024_octo, 2026_LBM}. However, due to the diversity of task goals and manipulated objects, it is often difficult to design explicit reward functions or leverage well-defined task priors for such settings. As a result, imitation learning (IL) has been widely adopted, enabling robots to acquire complex manipulation skills directly from expert demonstrations. In this paper, we primarily focus on general-purpose robotic manipulation through imitation learning. Robotic manipulation methods can be divided into standard action policies~\cite{2022_zero-shot-planners, 2023_peract, 2023_RPT, 2023_voxposer, 2024_3d-diffusion-policy, 2024_manigaussian, 2025_rekep, 2025_DIF, 2025_imanip, 2025_adapt-your-body, 2025_HA-Align, 2025_state-free-policy, 2025_know_your_camera, 2026_actionmap} and video action policies~\cite{2022_diffusers, 2023_daydreamer, 2024_GR1, 2023_UniPi, 2024_gr2, 2024_Seer, 2025_UVAM}. The former framework naively inputs instructions and visuals, outputting the actions to complete the tasks, while the latter pipeline predicts future videos together with action generation, with the claim that these world modeling abilities can help understand the task better and generate the actions more accurately. 

In recent years, with the triumph of large foundation models, integrating these tremendous models into robot learning to build Vision-Language-Action (VLA) Models has become a prominent trend~\cite{2024_vla}. This paradigm was pioneered by the groundbreaking RT-2~\cite{2023_rt2}, which formally introduced the concept of VLA. Subsequently, researchers across academia and industry have developed a diverse array of VLA models~\cite{2024_open-x-embodiedment, 2023_flamingo, 2024_cogcot, 2024_openvla, 2024_pi0, 2025_gemini-robotcs, 2025_nora, 2025_openvla-oft, 2025_smolvla, 2025_pi0.5, 2025_pi0.6, 2026_robovlms, 2026_pi0.7, 2026_xiaomi_robotics, 2026_qwen_robomanip, 2026_molmoact2, 2026_vlaconf}. These newer iterations address specific challenges, such as leveraging 3D spatial information~\cite{2024_3D-VLA, 2025_3D-CAVLA, 2025_4DVLA, 2025_spatialvla, 2026_GAM}, exploiting intermediate data (like subtask decomposition, future frame/video prediction or robot trajectory prediction)~\cite{2025_tracevla, 2025_cotvla, 2025_dreamvla, 2025_f1, 2025_flowvla, 2025_mm-act, 2025_molmoact, 2025_worldvla, 2025_rynnvla, 2025_univla, 2025_up-vla, 2025_reconvla, 2026_fastwam, 2026_dreamzero, 2026_kairos, 2026_zerowam, 2026_lingbot-va, 2026_imagewam, 2026_wam_ttt, 2026_lingbot_va_2, 2026_mva, 2026_flowwam} to enhance action generation, and designing post-training optimization methods, such as planning or reinforcement learning, to adapt to specific environments~\cite{2025_vla-reasoner, 2025_GRAPE, 2025_evolvevla, 2025_RIPT-VLA, 2025_simplevla, 2025_srpo, 2025_tgrpo, 2025_thinkact, 2025_vla-rl, 2025_world-env, 2025_ava-vla}. Additionally, a subset of VLAs explores some niche but important aspects~\cite{2025_chatvla, 2025_LLSA, 2025_emergence, 2025_memoryvla, 2025_mergevla, 2025_fast, 2025_vla0, 2026_vlm4vla, 2026_X-ICM, 2026_zeta, 2026_pion}, such as latent actions~\cite{2024_LAPA, 2025_motus, 2026_lara, 2026_dila, 2026_HiLAM}, lightweight VLAs~\cite{2025_tinyvla, 2025_cronusvla, 2025_FLOWER, 2025_vla-adapter} and VLAs in specific domains~\cite{2025_combatvla, 2025_gr00t-n1, 2025_humanoidvla, 2025_tactile}. 

Despite their different emphases, most VLA models follow a similar pipeline that builds on pretrained LLMs or VLMs to process visual observations and language instructions and derive action-relevant representations for policy learning. However, this shared pipeline admits numerous design choices spanning foundational components, perception, and action modeling. As a result, the rapidly evolving VLA landscape remains something of a ``primordial soup'': rich in ideas but lacking clear structure. Differences in model architectures, training protocols, and evaluation settings further make it difficult to determine which design choices truly matter.
To bring structure to this fragmented design space, we systematically reexamine VLA models under a unified framework and evaluation protocol. Through more than 500 controlled experiments, we study design choices across foundational components, perception essentials, and action modeling perspectives, and distill 12 key findings into a practical core recipe for building strong VLA models. These principles instantiate \textbf{VLANeXt}, a simple yet effective VLA that achieves state-of-the-art performance on LIBERO~\cite{2023_libero} and LIBERO-plus~\cite{2025_libero-plus}, while transferring effectively to real-world manipulation tasks.
Beyond the standard VLA formulation, we further examine whether the distilled recipe remains effective as the VLA paradigm evolves. We extend VLANeXt along four emerging directions: model scaling, latent-action pretraining, latent predictive representation learning, and world action modeling. These extensions form the \textbf{VLANeXt family}, allowing us to study whether the identified design principles generalize across different model scales and sources of prior. Our results show that the core recipe remains effective across these emerging paradigms, providing a common foundation for studying and developing evolving VLA models.

\section{Core Recipes for Building Strong VLA Models}
\label{sec:roadmap}

In this section, we detail the step-by-step evolution from a simple baseline to the base VLANeXt model, distilling the core recipe that later serves as a common foundation for the emerging-paradigm extensions. We organize our exploration along three aspects: foundational components (Sec.~\ref{sec:foundation}), perception essentials (Sec.~\ref{sec:perception}), and action modeling perspectives (Sec.~\ref{sec:perspective}). An overview is shown in Fig.~\ref{fig:roadmap}, with full results in Table~\ref{tab:roadmap}.

\textbf{Evaluation Setup.}  We perform the roadmap exploration on LIBERO and LIBERO-plus~\cite{2023_libero, 2025_libero-plus}. Main experiments are conducted on the spatial suite as our primary testbed, while the resulting insights generalize across the other suites (Object, Goal, and Long).

\textbf{Baseline.} Our baseline follows the VLA pipeline introduced in RT-2~\cite{2023_rt2}, the origin of VLAs, and later adopted by OpenVLA~\cite{2024_openvla}. We use LLaMA as the language backbone~\cite{2024_llama}. Since LLaMA does not natively support visual inputs, we also pair our backbone with SigLIP2~\cite{2025_siglip2} as the vision encoder.
A subset of rarely used text tokens is repurposed as action tokens, enabling action prediction in the same autoregressive framework. Continuous actions are discretized using a simple binning strategy and modeled as classification over bin indices.
We intentionally start from this minimal, classical RT-2-style~\cite{2023_rt2} setup to provide a clean reference point for analyzing the effects of different design choices. Our implementation adopts a more recent LLaMA version (LLaMA 3.2) but at a smaller scale (3B parameters), compared to OpenVLA~\cite{2024_openvla}.

\begin{figure}[t]
  \begin{center}
    \centerline{\includegraphics[width=0.93\columnwidth]{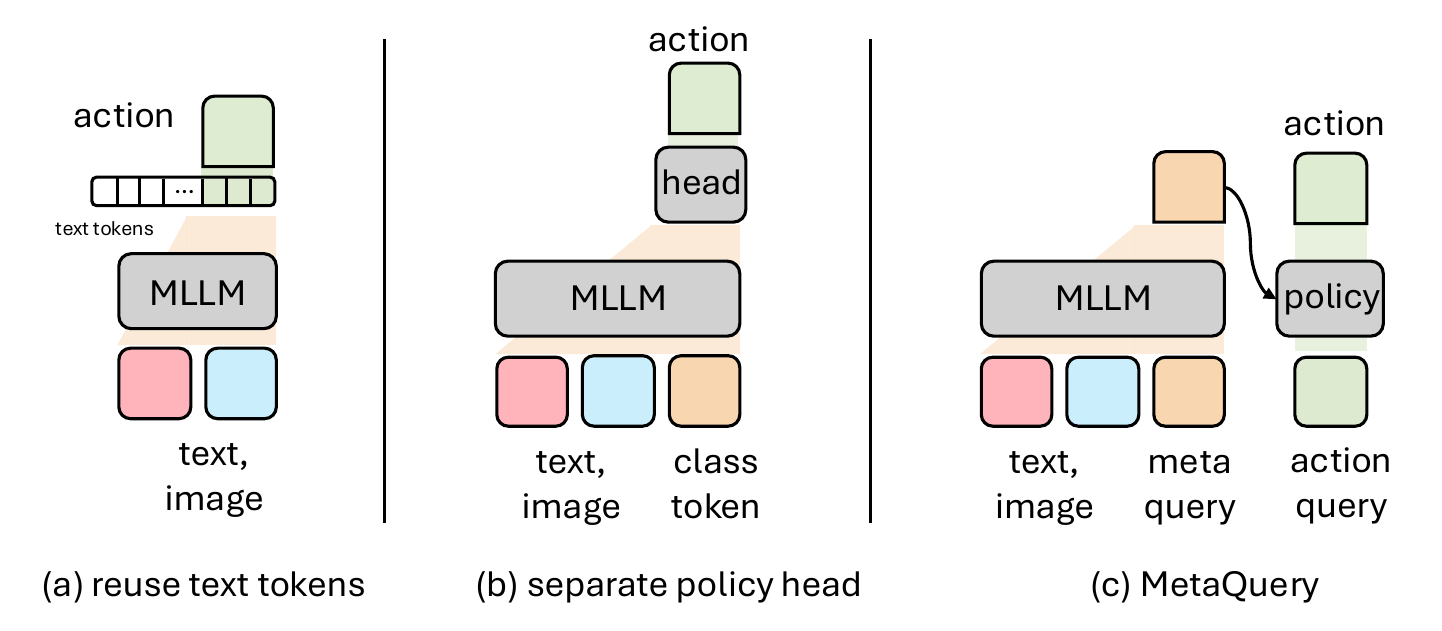}}
    \vspace{-2mm}
    \caption{
      Design choices for the policy module.
    }
    \label{fig:policy_module}
  \end{center}
  \vspace{-10mm}
\end{figure}

\subsection{The Foundational Components}
In this section, we investigate some core design choices of VLAs, including architectures and training losses. 

\label{sec:foundation}

\textbf{Policy Module Design.}
Our baseline follows RT-2~\cite{2023_rt2} and OpenVLA~\cite{2024_openvla}, reusing text tokens for action classification. We first examine whether an explicit policy head is necessary. To this end, we append a class token to the text and visual embeddings and feed its LLM output into a two-layer policy head (transformer architecture) for action classification (Fig.~\ref{fig:policy_module}(a)(b)). Results show that introducing a separate policy head performs slightly better than directly reusing text tokens (Table~\ref{tab:roadmap}), suggesting that \textit{decoupling action prediction from the linguistic token space is beneficial}.

We further investigate whether a more expressive policy module brings additional gains. Specifically, we replace the single class token with multiple tokens (16) and expand the policy network from 2 to 12 layers, making the design conceptually similar to MetaQuery~\cite{2025_metaquery} (Fig.~\ref{fig:policy_module}(c)). This\textit{ enlarged policy module yields a significant performance improvement} (Table~\ref{tab:roadmap}). The base VLANeXt adopts this design.

\textbf{Action Chunking.}
Our baseline predicts actions one step at a time. Here, we evaluate action chunking, which predicts multiple future actions jointly and is known to improve inference efficiency~\cite{2025_openvla-oft}. \textit{Results show that action chunking substantially improves over single-step prediction, while a horizon of 8 achieves the strongest robustness on LIBERO-plus} (Table~\ref{tab:roadmap}), suggesting that \textit{modeling a longer temporal window of action provides a more coherent view of the action sequence}. We therefore adopt action chunking with a chunk size of 8.

\textbf{Action Learning Objective.}
An action chunk is a continuous vector of shape $(t, dim)$. Our baseline discretizes this vector using binning (first normalizing to the range $[-1, 1]$, then dividing into 256 bins) and treats action prediction as classification, following OpenVLA~\cite{2024_openvla}. We compare this with alternative objectives, including direct regression~\cite{2025_openvla-oft}, diffusion-based losses such as DDIM~\cite{2021_ddim, 2025_dreamvla}, flow matching~\cite{2022_flow-matching, 2025_f1}, and VQ-VAE–based codebook (using a codebook of size 1,024, with three codes assigned to each action) classification~\cite{2017_vq-vae, 2021_vqgan}.

Results show that \textit{regression achieves the strongest performance}, \textit{with diffusion-based objectives close behind}, while classification-based approaches perform worst (Table~\ref{tab:roadmap}). In addition, as performance improves, the flow-matching objective eventually outperforms the regression loss, since the former can represent precise control signals. We therefore adopt the flow-matching objective. We also observe that classification using the VQ–VAE–based codebook underperforms relative to the binning strategy. We attribute this to the fact that the action spaces are low-dimensional, meaning a simple binning approach provides sufficient resolution.

\begin{figure}[t]
  \begin{center}
    \centerline{\includegraphics[width=\columnwidth]{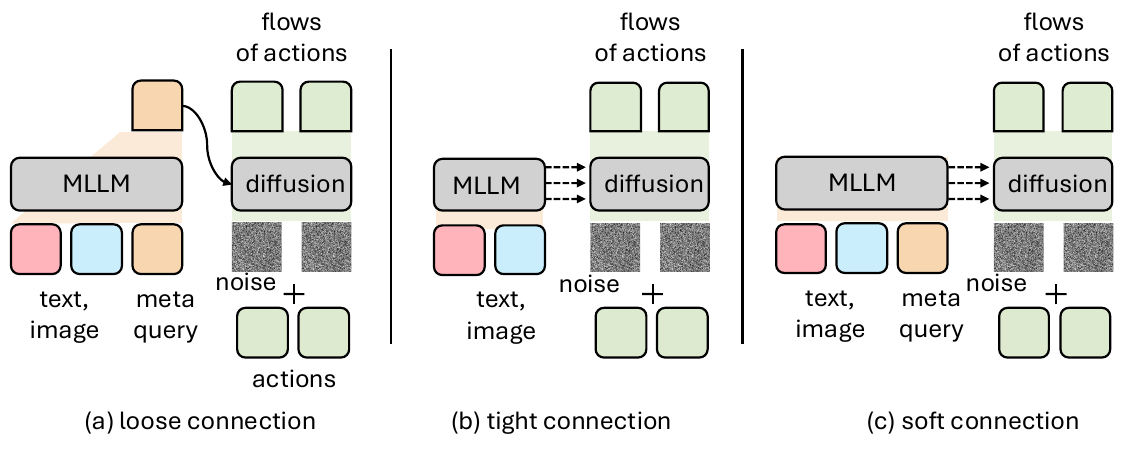}}
    \vspace{-2mm}
    \caption{
      Design choices for the VLM-Policy connection.
    }
    \label{fig:connection}
  \end{center}
  \vspace{-10mm}
\end{figure}

\textbf{VLM Backbone Capacity.}
Our baseline uses LLaMA as the backbone~\cite{2024_llama}. We evaluate alternative VLM backbones to study how backbone strength affects VLA performance, including PaliGemma-3B~\cite{2024_paligemma} (used in the $\pi$ series~\cite{2024_pi0, 2025_pi0.5}) and the Qwen-VL family~\cite{2025_qwen3-vl}, which represent some of the most capable open-source VLMs currently available.

Results show a consistent trend: \textit{more capable VLM backbones generally lead to stronger VLA performance in this controlled comparison} (Table~\ref{tab:roadmap}), with Qwen3-VL-4B outperforming Qwen3-VL-2B, which in turn outperforms LLaMA-3.2-3B and PaliGemma-3B. For the remainder of the core-recipe study, we use Qwen3-VL-2B as an effective and efficient choice, keeping backbone scale fixed while ablating other design components. We revisit model scaling after the full recipe is established in Sec. \ref{sec:new_paradigms} to examine whether the recipe remains effective across model scales.
This finding differs from that reported by Zhang et al.~\cite{2026_vlm4vla}. A possible reason is that our larger policy module can better exploit the representational capacity of stronger VLMs, whereas the lightweight policy head in Zhang et al.~\cite{2026_vlm4vla} may limit such gains. We leave a deeper investigation to future work.

\begin{table*}
    \centering
    \caption{Ablation across the VLA design space on LIBERO, LIBERO-plus (spatial suite). Each block varies in one design aspect. LIBERO-plus evaluates robustness under diverse perturbations. Performance improves steadily as effective design choices are incorporated.}
    \small
    \setlength{\tabcolsep}{4pt}
    \begin{tabular}{l | c | cccccccc}
\toprule
 & \textbf{LIBERO} & \multicolumn{8}{c}{\textbf{LIBERO-plus}} \\
 \midrule
\textbf{Model} & \textbf{Original} & \textbf{Camera} & \textbf{Robot} & \textbf{Language} & \textbf{Light} & \textbf{Background} & \textbf{Noise} & \textbf{Layout} & \textbf{Total} \\
\midrule
\rowcolor[HTML]{EFEFEF} 
\multicolumn{10}{l}{\textbf{\textit{Foundational Components}}} \\
\midrule
\rowcolor[HTML]{EFEFEF} 
\multicolumn{10}{l}{\textit{RT-2 like Baseline}} \\
Baseline     &   19.8  & -  & -  & - & -  & - & - & - & $<$ 5.0 \\
\midrule
\rowcolor[HTML]{EFEFEF}
\multicolumn{10}{l}{\textit{Policy Module Design}} \\
Baseline     &   19.8  & -  & -  & - & -  & - & - & - & $<$ 5.0 \\
Separate Head    &  30.2  & 0.8 & 10.0 & 31.0 & 15.4 & 24.8 & 4.0 & 30.1 & 16.6 \\
Large Policy Module  & 64.4  & 0.5 & 12.6 & 79.7 & 34.6 & 32.9 & 8.5 & 63.1 & 34.0 \\
\midrule
\rowcolor[HTML]{EFEFEF}
\multicolumn{10}{l}{\textit{Action Chunking Horizon}} \\
Action Chunk 1     &  64.4   & 0.5 & 12.6 & 79.7 & 34.6 & 32.9 & 8.5 & 63.1 & 34.0  \\
Action Chunk 4     &   75.4  & 5.3 & 28.0 & 67.9 & 42.5 & 50.4 & 14.8 & 70.4 & 40.0 \\
Action Chunk 8     &   74.6  & 5.6 & 26.0 & 85.6 & 56.8 & 55.8 & 11.7 & 63.9 & 43.4 \\
\midrule
\rowcolor[HTML]{EFEFEF}
\multicolumn{10}{l}{\textit{Action Learning Objective}} \\
bin Classification     &  74.6   & 5.6 & 26.0 & 85.6 & 56.8 & 55.8 & 11.7 & 63.9 & 43.4 \\
VQ-VAE Classification     &  58.8   & 3.2 & 44.3 & 67.2 & 42.5 & 43.8 & 7.1 & 48.3 & 36.5 \\
Regression     &   85.4  & 5.1 & 32.3 & 90.5 & 62.7 & 68.6 & 7.7 & 75.6 & 48.4  \\
DDIM     &   80.0  & 4.8 & 52.6 & 80.3 & 70.5 & 55.4 & 9.4 & 68.3 & 48.3 \\
Flow Matching     &   80.0  & 7.2  & 34.6 & 79.2 & 46.9 & 57.8 & 11.1 & 77.4 & 45.0 \\
\midrule
\rowcolor[HTML]{EFEFEF}
\multicolumn{10}{l}{\textit{VLM Backbone Capacity}} \\
PaliGemma     &   69.8  & 1.1 & 17.1 & 32.1 & 22.9 & 32.6 & 2.8 & 24.9 & 18.9 \\
LLaMA3.2 + SigLIP2     &  80.0  & 7.2  & 34.6 & 79.2 & 46.9 & 57.8 & 11.1 & 77.4 & 45.0 \\
Qwen3-VL-2B     &   90.0  & 9.6 & 42.0 & 74.6 & 75.0 & 68.6 & 27.9 & 83.6 & 53.7 \\
Qwen3-VL-4B     &   95.8  & 12.2 & 66.0 & 93.3 & 89.0 & 81.0 & 29.9 & 88.6 & 64.8 \\
\midrule
\rowcolor[HTML]{EFEFEF}
\multicolumn{10}{l}{\textit{VLM-Policy Connection}} \\
Loose Connection    &  90.0   & 9.6 & 42.0 & 74.6 & 75.0 & 68.6 & 27.9 & 83.6 & 53.7 \\
Tight Connection   &   90.0  & 14.4 & 51.7 & 81.0 & 68.5 & 67.8 & 25.1 & 82.1 & 55.4 \\
Soft Connection   &  91.8   & 11.8 & 58.3 & 89.2 & 72.9 & 74.4 & 19.9 & 72.5 & 56.2 \\
\midrule
\rowcolor[HTML]{EFEFEF} 
\multicolumn{10}{l}{\textbf{\textit{Perception Essentials}}} \\
\midrule
\rowcolor[HTML]{EFEFEF}
\multicolumn{10}{l}{\textit{Temporal Observation History}} \\
Current Frame Image    &  91.8   & 11.8 & 58.3 & 89.2 & 72.9 & 74.4 & 19.9 & 72.5 & 56.2\\
Temporal Observation History   &  85.0   & 7.2 & 68.6 & 51.5 & 65.8 & 62.0 & 20.8 & 80.8 & 50.2  \\
\midrule
\rowcolor[HTML]{EFEFEF}
\multicolumn{10}{l}{\textit{Camera View Horizon}} \\
Third-person camera view   &  91.8   & 11.8 & 58.3 & 89.2 & 72.9 & 74.4 & 19.9 & 72.5 & 56.2 \\
Multi-view (third-person + wrist)   & 97.6 & 64.9 & 54.0 & 91.8 & 97.7 & 93.0 & 85.5 & 90.1 & 80.5 \\
\midrule
\rowcolor[HTML]{EFEFEF}
\multicolumn{10}{l}{\textit{Proprioception Conditioning}} \\
No Proprioception Input  & 97.6 & 64.9 & 54.0 & 91.8 & 97.7 & 93.0 & 85.5 & 90.1 & 80.5 \\
Proprioception to VLM  & 98.0 & 87.2 & 62.2 & 86.2 & 98.3 & 93.8 & 92.0 & 96.6 & 87.7 \\
Proprioception to Policy  & 96.2 & 62.8 & 69.1 & 92.3 & 92.5 & 96.5 & 87.2 & 88.3 & 83.4  \\
Proprioception to VLM \& Policy  & 97.6 & 77.9 & 73.4 & 82.3 & 90.1 & 94.6 & 90.6 & 88.8 & 84.8 \\
\midrule
Linear Projector  & 98.0 & 87.2 & 62.2 & 86.2 & 98.3 & 93.8 & 92.0 & 96.6 & 87.7 \\
Transformer Projector  & 96.4 & 96.8 & 58.0 & 84.6 & 95.2 & 98.8 & 96.9 & 93.3 & 88.8 \\
Transformer Projector \& MAE  & 97.0 & 91.1 & 51.1 & 89.9 & 72.8 & 86.9 & 78.7 & 85.6 & 78.9 \\
\rowcolor[HTML]{EFEFEF} 
\midrule
\multicolumn{10}{l}{\textbf{\textit{Action Modeling Perspectives}}} \\
\midrule
\rowcolor[HTML]{EFEFEF}
\multicolumn{10}{l}{\textit{World Modeling Perspective}} \\
Normal  & 98.0 & 87.2 & 62.2 & 86.2 & 98.3 & 93.8 & 92.0 & 96.6 & 87.7 \\
World Modeling  & 98.0 & 94.4 & 80.3 & 76.9 & 99.7 & 98.8 & 93.2 & 93.2 & 90.3 \\
\midrule
\rowcolor[HTML]{EFEFEF}
\multicolumn{10}{l}{\textit{Time Series Forecasting Perspective}} \\
Normal  & 98.0 & 87.2 & 62.2 & 86.2 & 98.3 & 93.8 & 92.0 & 96.6 & 87.7 \\
Frequency Domain Loss  & 99.0 & 95.7 & 78.6 & 86.9 & 99.7 & 98.8 & 98.0 & 96.6 & 93.1 \\
\bottomrule
\end{tabular}
\label{tab:roadmap}
\end{table*}

\begin{figure*}[t]
  \begin{center}
    \centerline{\includegraphics[width=0.95\textwidth]{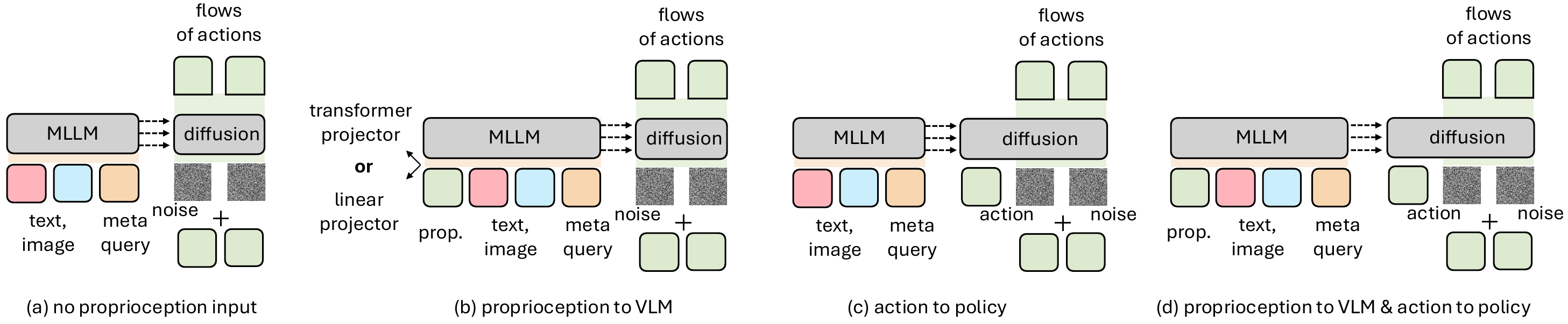}}
    \vspace{-1mm}
    \caption{
      Design choices for proprioception conditioning.
    }
    \label{fig:proprioception_condition}
  \end{center}
  \vspace{-10mm}
\end{figure*}

\textbf{VLM-Policy Connection.}
We next study how different connection strategies between the VLM and the policy module affect performance. Our baseline adopts a MetaQuery-style design~\cite{2025_metaquery}, as discussed in ``Policy Module Design''. We refer to this design as the loose strategy, where the VLM and policy module are fully decoupled. We compare this with a tight strategy that connects the two modules layer by layer, as in the $\pi$ series~\cite{2024_pi0, 2025_pi0.5}. Inspired by these two designs, we further introduce a soft strategy that also connects them layer by layer but inserts learnable queries as a latent buffer between the modules (Fig.~\ref{fig:connection}). For all three connection strategies, we use cross attention for conditioning, and the timestep is conditioned using adaLN, following Peebles and Xie~\cite{2023_dit}.

Results show that the \textit{soft strategy slightly outperforms both loose and tight connections} (Table~\ref{tab:roadmap}), suggesting that the learnable query buffer helps better transfer useful representations from the VLM's textual space to the policy module's action space. This may be viewed as introducing a latent buffer between the two components, analogous to reasoning in a latent space~\cite{2024_reasoning-latent-space}. We adopt the soft connection in subsequent models.

\subsection{The Perception Essentials}
\label{sec:perception}

In this section, we shift our focus from foundational components to perception, investigating whether and how different modalities (\textit{e.g.}, visual observations and proprioceptive states) should be provided as inputs to VLAs.

\textbf{Temporal Observation History.}
We examine whether incorporating temporal observation history improves performance. Our baseline follows OpenVLA~\cite{2024_openvla} and uses only the current frame as input. We extend this to include multiple past frames, leveraging the video capability of the Qwen3-VL-2B backbone~\cite{2025_qwen3-vl} for a controlled comparison.
Results show that \textit{simply providing additional past observations as VLM inputs does not improve action generation} and slightly degrades performance under the standard VLA formulation (Table~\ref{tab:roadmap}), indicating that redundant temporal inputs may introduce noise or distract the model, even though the backbone is already pretrained for video understanding. However, explicitly modeling temporal dynamics may be more effective, which we revisit through predictive representation learning and world action modeling in Sec. \ref{sec:new_paradigms}.

\textbf{Camera View Horizon.}
We study the effect of camera viewpoints on VLA performance. Our baseline uses a single third-person view, following OpenVLA~\cite{2024_openvla}. Many robotics datasets~\cite{2024_open-x-embodiedment, 2024_droid} additionally provide an in-hand wrist camera, allowing choices between single-view and multi-view inputs.
Results show that \textit{combining third-person and wrist views significantly improves performance} (Table~\ref{tab:roadmap}), suggesting that multi-view observations provide complementary geometric cues that help resolve spatial ambiguities.

\textbf{Proprioception Conditioning.}
We examine the role of proprioception, which provides information about the robot’s internal state and motion history. Our baseline, following OpenVLA~\cite{2024_openvla}, does not use proprioceptive inputs. We compare three variants: conditioning the VLM, conditioning the policy module, and conditioning both (Fig.~\ref{fig:proprioception_condition}). Specifically, proprioception is provided as input to the VLM. For the policy, we use historical actions as input to align with the generated actions.

Results show that \textit{conditioning proprioception in the VLM yields the best performance} (Table~\ref{tab:roadmap}). We hypothesize that integrating proprioception at the VLM level allows better fusion with visual and language inputs, whereas injecting it directly into the policy module may reduce reliance of action prediction on visual observations and instructions.
Although this appears to differ from the conclusion reported by Zhao \textit{et al.}~\cite{2025_state-free-policy}, who claim that proprioception is not needed, their study evaluates architectures where proprioception is injected only into the policy module. In that setting, removing proprioception improves performance, which is consistent with our LIBERO results.

We further compare three different integration mechanisms, including a linear projector, a transformer-based projector, and a transformer projector with masked reconstruction pretraining~\cite{2022_mae}. The \textit{transformer-based projector performs slightly better on LIBERO-plus} (Table~\ref{tab:roadmap}); for simplicity, we use the linear projector in the core recipe of our base design.

\begin{figure}[t]
  \begin{center}
    \centerline{\includegraphics[width=\columnwidth]{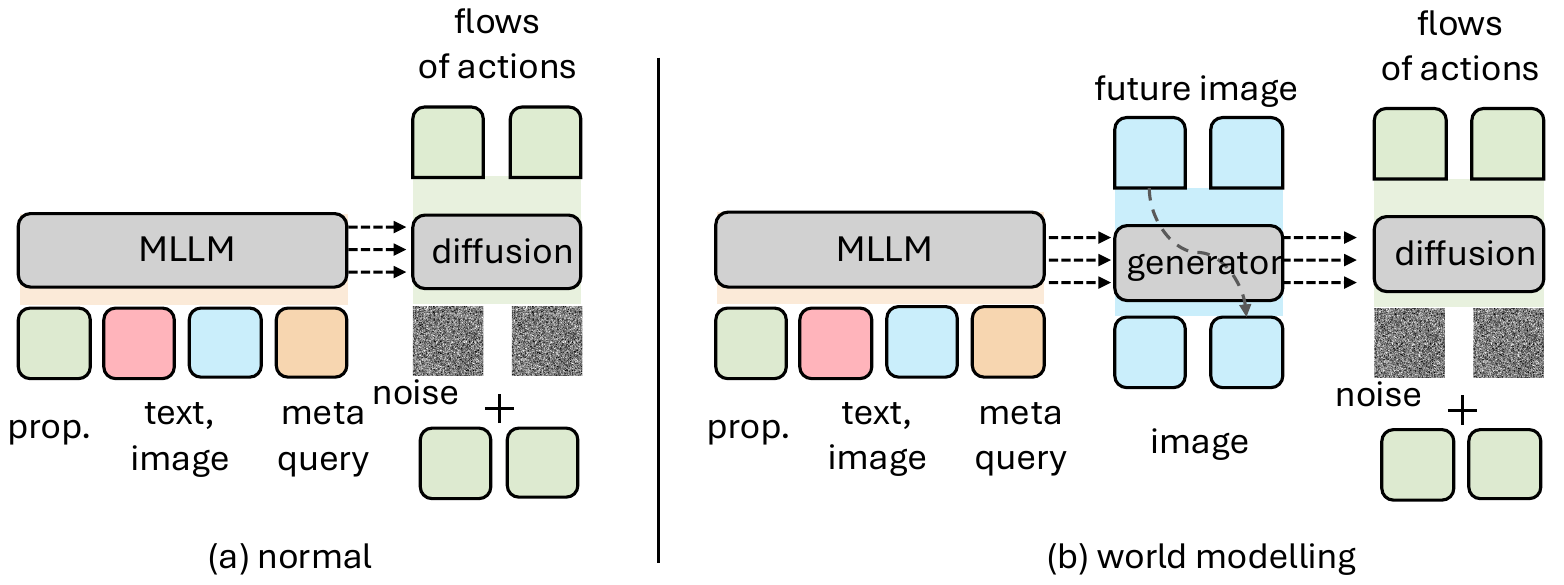}}
    \vspace{-1mm}
    \caption{
      Augmenting action prediction with an auxiliary world modeling objective.
    }
    \label{fig:world_modeling}
  \end{center}
  \vspace{-8mm}
\end{figure}

\subsection{Action Modeling Perspectives}
\label{sec:perspective}
Here, we examine auxiliary design and training objectives to facilitate action generation.

\textbf{World Modeling.}
We first study world modeling as an auxiliary objective within the standard VLA architecture~\cite{2025_f1, 2025_worldvla}. To isolate the effect of dynamics supervision, we deliberately avoid using a pretrained visual generator. Instead, we tokenize images using the Emu3.5 image tokenizer~\cite{2025_emu3.5} and predict future image tokens with a next-token objective. The target is the future frame at a fixed horizon (8 steps, aligned with the action chunk length). The visual generation module is inserted between the VLM and the policy module with layer-wise connections (Fig.~\ref{fig:world_modeling}).
\textit{Adding world modeling improves action generation performance} (Table~\ref{tab:roadmap}), indicating that predicting future observations is beneficial. Although this auxiliary world-modeling objective improves action prediction, it nearly triples the training cost. We therefore exclude this auxiliary branch from the core recipe of the base VLANeXt. Later, in Sec. \ref{sec:new_paradigms}, we revisit explicit future prediction through two emerging paradigms: JEPA-style latent predictive representation learning and world action modeling with a pretrained video-generation backbone.

\textbf{Time Series Forecasting.}
We also explore facilitating action generation from a time-series forecasting perspective. Inspired by frequency-domain modeling in time-series prediction~\cite{2022_fedformer, 2023_frequency-mlp, 2024_rethinking-fourier, 2025_fredf}, we introduce a simple auxiliary loss that minimizes the MSE between predicted and ground-truth actions in the frequency domain. We use the discrete cosine transform~\cite{1974_dct} to convert the action to the frequency domain, and assign higher weights to low-frequency components and lower weights to high-frequency components, as high-frequency components are often noisier.

This strategy \textit{improves action generation performance}, slightly surpassing the auxiliary world-modeling variant studied above while adding negligible training overhead (Table~\ref{tab:roadmap}). The gain likely arises because it serves as a regularization term to prevent the model from overfitting to trajectory jitter, thereby improving generalization.

\begin{figure}[t]
  \begin{center}
    \centerline{\includegraphics[width=\columnwidth]{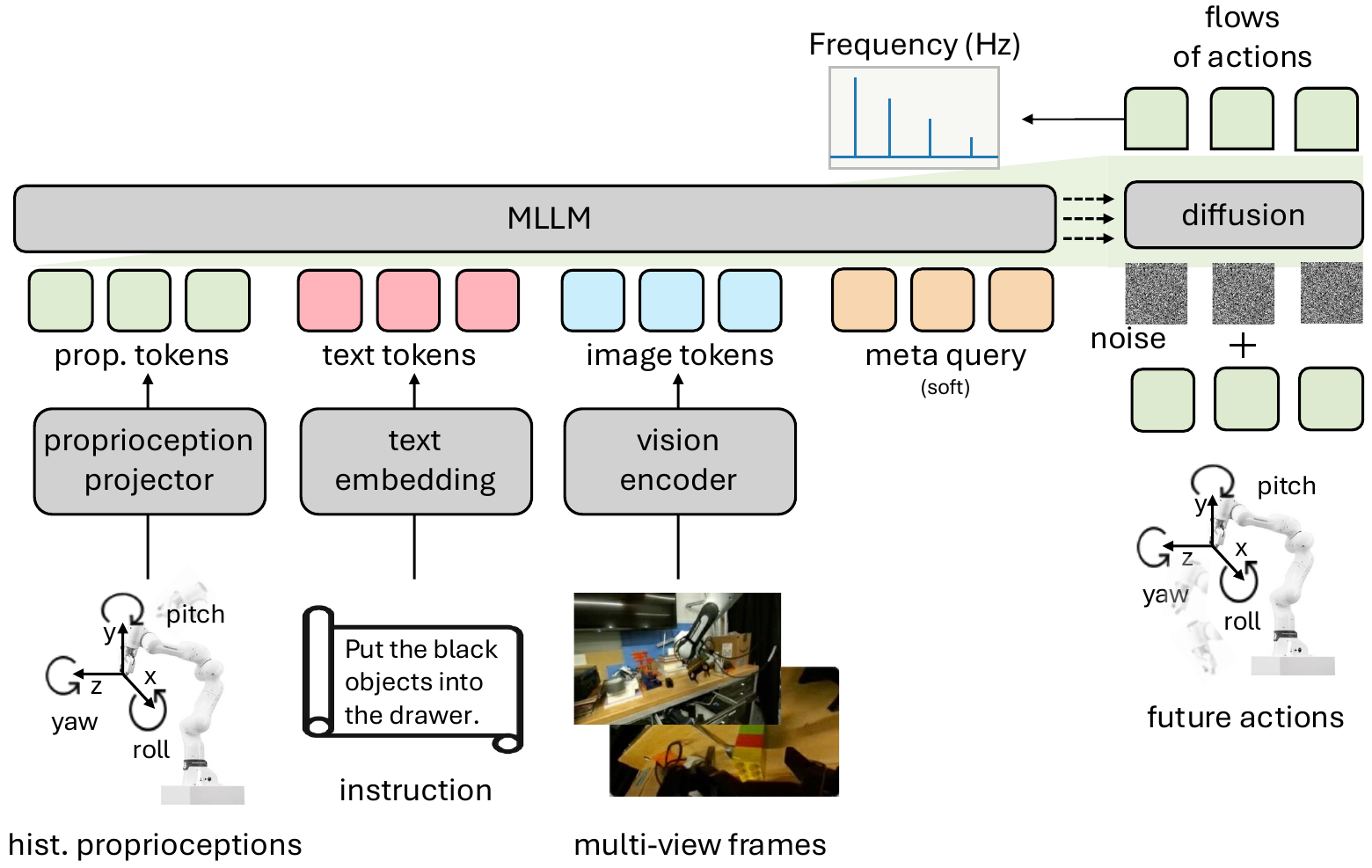}}
    \vspace{-1mm}
    \caption{
      \textbf{VLANeXt architecture}. Multi-view visual inputs, language instructions, and proprioception are tokenized and processed by a multimodal LLM, with meta queries enabling soft interaction with the policy module. Action chunks are predicted using flow matching and further regularized by a frequency-domain objective.
    }
    \label{fig:framework}
  \end{center}
  \vspace{-8mm}
\end{figure}

\subsection{Summary of Recipes}
Starting from a classical RT-2/OpenVLA-style baseline, we find that strong VLA performance emerges from a series of principled design choices. Beneficial changes include: replacing token reuse with a deeper, dedicated policy module; adopting action chunking to model longer temporal action horizons; using continuous objectives such as flow matching (with regression also effective under simple distributions); employing a stronger VLM backbone (Qwen3-VL-2B as an effective–efficient choice); and connecting the VLM and policy module through soft, layer-wise interactions with learnable query buffers. 

On the perception side, multi-view inputs (third-person + wrist) and VLM-side proprioception conditioning improve performance, while simply adding past observations as additional VLM inputs does not improve performance under the standard formulation. Moreover, adding a lightweight frequency-domain auxiliary loss further boosts action generation with negligible cost. Although the auxiliary world-modeling variant studied above also improves performance, its substantially higher training cost makes it less practical. Together, these choices define the \textbf{core recipe} and instantiate the base VLANeXt model. Importantly, the recipe above is distilled within the standard VLA formulation. In Secs. \ref{sec:new_paradigms} and \ref{sec:new_paradigms_evaluation}, we further examine whether these principles remain effective as VLAs scale and evolve toward latent-action, predictive-representation, and world-action modeling paradigms.

\section{Evaluation of the Core Recipe}
\label{sec:core_recipe_evaluation}

\subsection{Benchmark Evaluations}
\subsubsection{Settings}
Having distilled the core recipe, we first evaluate the resulting base VLANeXt under the standard VLA formulation, covering both simulation benchmarks and real-world deployment. The emerging-paradigm extensions are evaluated separately in Sec. \ref{sec:new_paradigms_evaluation}. To evaluate both standard performance and generalization robustness, we employ the LIBERO ecosystem. We first evaluate our base VLANeXt on the standard LIBERO benchmark~\cite{2023_libero}, which provides four distinct categories (Spatial, Object, Goal, and Long) to test the task learning ability, each providing 500 expert episodes across 10 tasks to assess policy performance across different spatial layouts, objects, goals, and long-horizon tasks. 

To test the generalization boundaries of our model further, we evaluate our method on LIBERO-plus~\cite{2025_libero-plus}. Unlike the static conditions in standard LIBERO, LIBERO-plus introduces systematic variations to the evaluation episodes, comprising 10,030 episodes across the above four suites in LIBERO, with perturbations in visual (\textit{e.g.}, lighting, background, camera pose), physical (\textit{e.g.}, object layout, robot state), and semantic (\textit{e.g.}, language instruction rewrites) dimensions. 

Following the standard setting in OpenVLA~\cite{2024_openvla}, we train our models on the modified LIBERO dataset for each suite (Spatial, Object, Goal, and Long), and evaluate performance on both the LIBERO and LIBERO-plus benchmarks (which include unseen perturbations) for the corresponding suite. Notably, for LIBERO-plus, models are trained only on the original LIBERO data and directly evaluated under unseen perturbations, without perturbation-specific training. For fair comparisons across different design choices, we directly fine-tune all models on the LIBERO dataset. All core-recipe ablations in Sec. \ref{sec:roadmap} use 10,000 training steps with a batch size of 256. The learning rate is set to $1\times10^{-4}$ for models smaller than 3B parameters and $5\times10^{-5}$ otherwise in our explorations.

\subsubsection{LIBERO Benchmark Results}
On the LIBERO benchmark, we compare our method against two categories of approaches: (i) direct policy learning methods that are trained solely on robotic datasets, and (ii) VLA methods that leverage knowledge from pretrained VLMs for policy learning. For direct policy learning methods, we include Diffusion Policy~\cite{2025_diffusion-policy}, Octo~\cite{2024_octo}, and MDT~\cite{2024_MDT}. For VLA methods, we compare against OpenVLA~\cite{2024_openvla}, TraceVLA~\cite{2025_tracevla}, SpatialVLA~\cite{2025_spatialvla}, WorldVLA~\cite{2025_worldvla}, CoT-VLA~\cite{2025_cotvla}, $\pi_0$~\cite{2024_pi0}, $\pi_0$-Fast~\cite{2025_fast}, NORA~\cite{2025_nora}, SmolVLA~\cite{2025_smolvla}, UniVLA~\cite{2025_univla}, FLOWER~\cite{2025_FLOWER}, and OpenVLA-OFT~\cite{2025_openvla-oft}.

The comparison results are shown in Table \ref{tab:libero}. As we can observe, the base VLANeXt instantiated by the core recipe achieves state-of-the-art performance, demonstrating that the distilled design principles provide a strong foundation for the standard VLA formulation.

\subsubsection{LIBERO-plus Benchmark Results}
For the LIBERO-plus benchmark, we compare our model with several VLA models such as OpenVLA~\cite{2024_openvla}, WorldVLA~\cite{2025_worldvla}, NORA~\cite{2025_nora}, UniVLA~\cite{2025_univla}, $\pi_0$~\cite{2024_pi0},  $\pi_0$-Fast~\cite{2025_fast}, and OpenVLA-OFT~\cite{2025_openvla-oft}. 

As shown in Table \ref{tab:libero-plus}, the proposed VLANeXt model demonstrates strong generalization ability across different types of unseen perturbations. 
Moreover, VLANeXt improves the success rate by 14.3 percentage points over the previous state-of-the-art method, OpenVLA-OFT~\cite{2025_openvla-oft}, on LIBERO-plus, suggesting that the core recipe also provides strong robustness and generalization under unseen perturbations.

\begin{table}[t]
\centering
\caption{LIBERO benchmark performance. The results are shown in success rate (\%). S, O, G, L: Spatial, Object, Goal, and Long suites, respectively. We color the \colorbox{tabfirst}{\textbf{best}} and \colorbox{tabsecond}{second-best} results.}
\small
\setlength{\tabcolsep}{8pt}
\begin{tabular}{l | cccc | c}
\toprule
\textbf{Model} & \textbf{S} & \textbf{O} & \textbf{G} & \textbf{L} & \textbf{Avg} \\
\midrule
\rowcolor[HTML]{EFEFEF} 
\multicolumn{6}{l}{\textit{Baseline Direct Policy Models}} \\
Diffusion Policy     & 78.3 & 92.5 & 68.3 & 50.5 & 72.4 \\
Octo     & 78.9 & 85.7 & 84.6 & 51.5 & 75.1 \\
MDT     & 78.5 & 87.5 & 73.5 & 64.8 & 76.1 \\
\midrule
\rowcolor[HTML]{EFEFEF} 
\multicolumn{6}{l}{\textit{Baseline VLA Models}} \\
TraceVLA     & 84.6 & 85.2 & 75.1 & 54.1 & 74.8 \\
OpenVLA     & 84.7 & 88.4 & 79.2 & 53.7 & 76.5 \\
SpatialVLA     & 88.2 & 89.9 & 78.6 & 55.5 & 78.1 \\
WorldVLA     & 85.6 & 89.0 & 82.6 & 59.0 & 79.1 \\
CoT-VLA     & 87.5 & 91.6 & 87.6 & 69.0 & 83.9 \\
$\pi_0$-Fast     & 96.4 & 96.8 & 88.6 & 60.2 & 85.5 \\
$\pi_0$     & 90.0 & 86.0 & 95.0 & 73.0 & 86.0 \\
NORA     & 92.2 & 95.4 & 89.4 & 74.6 & 87.9 \\
SmolVLA     & 93.0 & 94.0 & 91.0 & 77.0 & 88.8 \\
UniVLA     & 96.5 & 96.8 & 95.6 & 92.0 & 95.2 \\
FLOWER     & 97.5 & \csecond{99.1} & 96.1 & \cfirst{94.9} & 96.9 \\
$\pi_{0.5}$     & \csecond{98.8} & 98.2 & \cfirst{98} & 92.4 & 96.9 \\
OpenVLA-OFT     & 97.6 & {98.4} & \csecond{97.9} & 94.5 & \csecond{97.1} \\
\midrule
\rowcolor[HTML]{EFEFEF} 
\multicolumn{6}{l}{\textit{Ours}} \\
\textbf{VLANeXt} & \cfirst{99.0} & \cfirst{99.2} & 96.6 & \csecond{94.8} & \cfirst{97.4} \\
\bottomrule
\end{tabular}
\label{tab:libero}
\vspace{-3mm}
\end{table}

\begin{table*}[t]
\centering
\caption{LIBERO-plus benchmark performance. All the evaluated methods are trained using the LIBERO dataset and directly tested on the LIBERO-plus benchmark with unseen perturbations. The results are shown in success rate (\%). We color the \colorbox{tabfirst}{\textbf{best}} and \colorbox{tabsecond}{second-best} results on average. The complete per-suite results of the listed methods can also be found in the LIBERO-plus study by Fei et al.~\cite{2025_libero-plus}.}
\small
\setlength{\tabcolsep}{8pt}
\begin{tabular}{l | c | ccccccc | c}
\toprule
\textbf{Model} & \textbf{Suite} & \textbf{Camera} & \textbf{Robot} & \textbf{Language} & \textbf{Light} & \textbf{Background} & \textbf{Noise} & \textbf{Layout} & \textbf{Total} \\
\midrule
\rowcolor[HTML]{EFEFEF} 
\multicolumn{10}{l}{\textit{Baseline VLA Models}} \\
OpenVLA      & Average  & 0.8  & 3.5  & 23.0 & 8.1  & 34.8 & 15.2 & 28.5 & 15.6 \\
WorldVLA     & Average  & 0.1  & 27.9 & 41.6 & 43.7 & 17.1 & 10.9 & 38.0 & 25.0 \\
NORA         & Average  & 2.2  & 37.0 & 65.1 & 45.7 & 58.6 & 12.8 & 62.1 & 39.0 \\
UniVLA       & Average  & 1.8  & \csecond{46.2} & 69.6 & 69.0 & 81.0 & 21.2 & 31.9 & 42.9 \\
$\pi_0$      & Average  & 13.8 & 6.0  & 58.8 & 85.0 & 81.4 & \csecond{79.0} & 68.9 & 53.6 \\
$\pi_0$-Fast  & Average & \csecond{65.1} & 21.6 & 61.0 & 73.2 & 73.2 & 74.4 & 68.8 & 61.6 \\
\midrule
\multirow{5}{*}{OpenVLA-OFT}&Spatial & 88.3 & 40.0 & 80.5 & 98.3 & 97.3 & 96.3 & 93.9 & 84.0 \\
&Object & 38.9 & 25.4 & 99.0 & 73.7 & 97.6 & 72.3 & 71.8 & 66.5 \\
&Goal    & 62.0 & 25.2 & 53.2 & 93.9 & 92.5 & 75.2 & 59.1 & 63.0 \\
&Long    & 38.7 & 38.2 & 87.0 & 89.4 & 86.8 & 63.5 & 76.9 & 66.4 \\
& Average & 56.4 & 31.9 & \csecond{79.5} & \csecond{88.7} & \cfirst{93.3} & 75.8 & \csecond{74.2} & \csecond{69.6} \\
\midrule
\rowcolor[HTML]{EFEFEF}
\multicolumn{10}{l}{\textit{Ours}} \\

\multirow{5}{*}{VLANeXt} &Spatial   & 95.7 & 78.6 & 86.9 & 99.7 & 98.8 & 98.0 & 96.6 & 93.1 \\
&Object   & 99.5 & 48.5 & 98.6 & 99.3 & 84.7 & 99.8 & 78.2 & 86.5 \\
&Goal   & 96.6 & 63.6 & 51.5 & 97.5 & 70.8 & 96.8 & 63.9 & 76.2 \\
&Long   & 69.7 & 72.0 & 90.3 & 86.9 & 75.8 & 81.7 & 84.3 & 79.7 \\
  &  Average  & \cfirst{90.4} & \cfirst{65.7} & \cfirst{81.8} & \cfirst{95.9} & \csecond{82.5} & \cfirst{94.1} & \cfirst{80.8} & \cfirst{83.9} \\
\bottomrule
\end{tabular}
\label{tab:libero-plus}
\vspace{-5mm}
\end{table*}

\subsection{Real-World Evaluations}
To comprehensively assess the performance of our method, we additionally evaluate it in real-world deployments.

\begin{figure}[t]
  \begin{center}
    \centerline{\includegraphics[width=0.9\columnwidth]{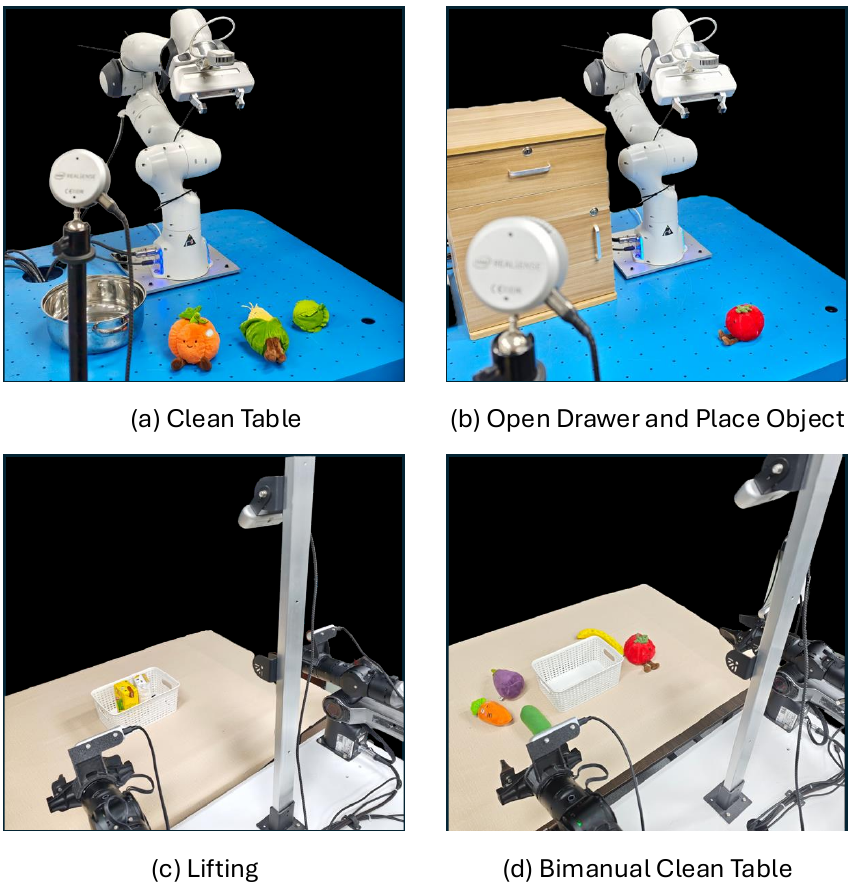}}
    \vspace{-2mm}
    \caption{
      Our single-arm and bimanual arm tasks for the real-world experiments. 
    }
    \label{fig:real_world_setting}
  \end{center}
  \vspace{-8mm}
\end{figure}

\subsubsection{Settings}
We design four tasks, including two single-arm tasks and two bimanual tasks, to evaluate our method. The single-arm tasks include table cleaning, which involves picking up objects from a table and placing them into a container, and drawer manipulation, where the robot opens a drawer, places objects inside, and closes it. The bimanual tasks include basket lifting, which requires lifting a basket using both hands, and bimanual table cleaning, where two arms coordinate to collect objects from a table and place them into a container. The single-arm experiments use Franka Emika, while the bimanual experiments are conducted on the Aloha system~\cite{2023_aloha}. A visualization of the experimental setup for each task is depicted in Figure \ref{fig:real_world_setting}.

For training, we collect 50 episodes per task and evaluate each model over 20 trials, reporting the success rate. We first pretrain the model on the DROID dataset~\cite{2024_droid} for 100k steps, then fine-tune it on each task for 20k steps with a learning rate of $1\times10^{-4}$. Because DROID contains only single-arm data, adapting the model to bimanual tasks requires reinitializing the proprioception projector and the final layer of the action generation module, while keeping all other pretrained weights.

\subsubsection{Results}
We compare against two representative VLA baselines, OpenVLA-OFT~\cite{2025_openvla-oft} and $\pi_0$~\cite{2024_pi0}. We load their pretrained checkpoints and fine-tune them on each task in the same manner as our method to ensure a fair comparison. The results are shown in Table \ref{tab:real_world_results}. As can be seen, our model performs well in real-world experiments, demonstrating that the core recipe yields a strong base VLA that performs effectively in real-world deployment. Moreover, despite being pretrained only on single-arm DROID data, our method can adapt to bimanual tasks after task-specific fine-tuning, demonstrating its adaptability across embodiments. Additional video demonstrations of our experimental results are provided in the supplementary materials.

Together, these results validate the core recipe under the standard VLA formulation, from simulated benchmarks to real-world manipulation. We next examine whether the same design principles remain effective as the VLA paradigm itself evolves.

\begin{table}[t]
\centering
\caption{Real-world evaluation results. Results are shown as (success count/total count). We color the \colorbox{tabfirst}{\textbf{best}} and \colorbox{tabsecond}{second-best} results.}
\small
\setlength{\tabcolsep}{8pt}
\begin{tabular}{l | cccc}
\toprule
\rowcolor[HTML]{EFEFEF} 
& \multicolumn{2}{c}{\textbf{Single Arm}} & \multicolumn{2}{c}{\textbf{Bimanual Arm}}\\
\textbf{Model} & \textbf{Clean} & \textbf{Drawer} & \textbf{Clean} & \textbf{Lifting} \\
\midrule
\rowcolor[HTML]{EFEFEF} 
\multicolumn{5}{l}{\textit{Baseline VLA Models}} \\
OpenVLA-OFT     & 7/20 & 7/20 & 5/20 & 9/20   \\
$\pi_0$     & \csecond{10/20} & \csecond{8/20} & \csecond{10/20} &  \csecond{13/20}   \\
\midrule
\rowcolor[HTML]{EFEFEF} 
\multicolumn{5}{l}{\textit{Ours}} \\
\textbf{VLANeXt} & \cfirst{14/20} & \cfirst{11/20}& \cfirst{11/20} & \cfirst{15/20}  \\
\bottomrule
\end{tabular}
\label{tab:real_world_results}
\vspace{-2mm}
\end{table}

\begin{table*}[t]
\centering
\small
\caption{
Ablation of the core recipe across different emerging paradigms in VLANeXt on the LIBERO benchmark. The table shows the average performance across the four suites of the LIBERO benchmark.
}
\begin{tabular}{l|l|cccccc}
\toprule
\textbf{Model}
& \textbf{Backbone}
& \textbf{Full}
& \textbf{w/o Chunk}
& \textbf{w/o Multi-view}
& \textbf{w/o Flow}
& \textbf{w/o Prop.}
& \textbf{w/o Freq.} \\
\midrule
VLANeXt-S
& Qwen3.5-0.8B
& 96.7
& 88.7 {\scriptsize\textcolor{dropcolor}{(-8.0)}}
& 89.9 {\scriptsize\textcolor{dropcolor}{(-6.8)}}
& 94.0 {\scriptsize\textcolor{dropcolor}{(-2.7)}}
& 95.8 {\scriptsize\textcolor{dropcolor}{(-0.9)}}
& 96.6 {\scriptsize\textcolor{dropcolor}{(-0.1)}} \\

VLANeXt-L
& Qwen3-VL-4B
& 98.4
& 93.2 {\scriptsize\textcolor{dropcolor}{(-5.2)}}
& 94.6 {\scriptsize\textcolor{dropcolor}{(-3.8)}}
& 96.6 {\scriptsize\textcolor{dropcolor}{(-1.8)}}
& 96.8 {\scriptsize\textcolor{dropcolor}{(-1.6)}}
& 97.1 {\scriptsize\textcolor{dropcolor}{(-1.3)}} \\

VLANeXt-LAM
& Qwen3-VL-2B
& 97.6
& 87.8 {\scriptsize\textcolor{dropcolor}{(-9.8)}}
& 85.8 {\scriptsize\textcolor{dropcolor}{(-11.8)}}
& 93.2 {\scriptsize\textcolor{dropcolor}{(-4.4)}}
& 94.6 {\scriptsize\textcolor{dropcolor}{(-3.0)}}
& 95.3 {\scriptsize\textcolor{dropcolor}{(-2.3)}} \\

VLANeXt-JEPA
& Qwen3-VL-2B
& 97.7
& 91.2 {\scriptsize\textcolor{dropcolor}{(-6.5)}}
& 93.3 {\scriptsize\textcolor{dropcolor}{(-4.4)}}
& 96.7 {\scriptsize\textcolor{dropcolor}{(-1.0)}}
& 95.9 {\scriptsize\textcolor{dropcolor}{(-1.8)}}
& 95.5 {\scriptsize\textcolor{dropcolor}{(-2.2)}} \\

VLANeXt-WAM
& WAN2.2-5B
& 98.2
& 94.1 {\scriptsize\textcolor{dropcolor}{(-4.1)}}
& 89.9 {\scriptsize\textcolor{dropcolor}{(-8.3)}}
& 96.0 {\scriptsize\textcolor{dropcolor}{(-2.2)}}
& 97.6 {\scriptsize\textcolor{dropcolor}{(-0.6)}}
& 98.0 {\scriptsize\textcolor{dropcolor}{(-0.2)}} \\
\bottomrule
\end{tabular}
\label{tab:core_recipe_generalization}
\vspace{-2mm}
\end{table*}

\begin{table*}[t]
\centering
\small
\caption{
Comparison of latent-action modeling variants on the LIBERO benchmark.
}
\begin{tabular}{l|l|cc|ccccc}
\toprule
\multirow{2}{*}{\textbf{Model}}
& \multirow{2}{*}{\textbf{Pretraining}}
& \multicolumn{2}{c|}{\textbf{Training Steps}}
& \multicolumn{5}{c}{\textbf{LIBERO}} \\
\cmidrule(lr){3-4}
\cmidrule(lr){5-9}
& & Pretrain & Finetune & Spatial & Object & Goal & Long & Avg \\
\midrule
\rowcolor[HTML]{EFEFEF}
\multicolumn{9}{l}{\textit{Baseline VLA Models}} \\
VLANeXt
& --
& -- & 10k
& 99.0 & 99.2 & 96.6 & 94.8 & 97.4 \\
VLANeXt (matched training duration)
& --
& -- & 30k
& 98.8 & 98.8 & 96.2 & 94.0 & 97.0 \\
\midrule
\rowcolor[HTML]{EFEFEF}
\multicolumn{9}{l}{\textit{Latent Action Modeling}} \\
VLANeXt-LAM
& LAM (VICReg)
& 20k & 10k
& 95.4 & 99.6 & 97.0 & 95.4 & 96.9 \\
VLANeXt-LAM
& LAM (VAE)
& 20k & 10k
& 98.0 & 97.8 & 95.2 & 96.2 & 96.8 \\
VLANeXt-LAM
& LAM (VQ-VAE)
& 20k & 10k
& 98.0 & 99.8 & 95.6 & 97.0 & 97.6 \\
\bottomrule
\end{tabular}
\label{tab:lam_comparison}
\vspace{-2mm}
\end{table*}

\begin{table*}[t]
\centering
\small
\caption{
Ablation of latent predictive objectives in VLANeXt-JEPA on the LIBERO benchmark.
}
\begin{tabular}{l|l|l|ccccc}
\toprule
\textbf{Model}
& \textbf{Generator}
& \textbf{Objective}
& \multicolumn{5}{c}{\textbf{LIBERO}} \\
\cmidrule(lr){4-8}
& & & Spatial & Object & Goal & Long & Avg \\
\midrule
\rowcolor[HTML]{EFEFEF}
\multicolumn{8}{l}{\textit{Baseline VLA Model}} \\
VLANeXt
& --
& action only
& 99.0 & 99.2 & 96.6 & 94.8 & 97.4 \\
\midrule
\rowcolor[HTML]{EFEFEF}
\multicolumn{8}{l}{\textit{Latent Predictive Representation Learning}} \\
VLANeXt
& Emu3.5 + transformer
& action + final image of current horizon
& 97.6 & 99.8 & 95.4 & 92.8 & 96.4 \\
VLANeXt-JEPA
& DINOv3 encoder + DiT
& action + final image of the episode
& 98.0 & 99.2 & 96.4 & 93.0 & 96.6 \\
VLANeXt-JEPA
& DINOv3 encoder + DiT
& action + final image of current horizon
& 98.0 & 99.8 & 97.0 & 95.8 & 97.7 \\
\bottomrule
\end{tabular}
\label{tab:jepa_comparison}
\vspace{-2mm}
\end{table*}

\begin{table*}[t]
\centering
\small
\caption{
Comparison of world action modeling variants on the LIBERO benchmark.
The action-only baseline uses the same WAN2.2-5B backbone and training setup
but removes the video-generation objective.
}
\begin{tabular}{l|l|l|l|ccccc}
\toprule
\textbf{Model}
& \textbf{Backbone}
& \textbf{Objective}
& \textbf{Video \& Action Connection}
& \multicolumn{5}{c}{\textbf{LIBERO}} \\
\cmidrule(lr){5-9}
& & & & Spatial & Object & Goal & Long & Avg \\
\midrule
\rowcolor[HTML]{EFEFEF}
\multicolumn{9}{l}{\textit{Baseline VLA Models}} \\
VLANeXt
& Qwen3-VL-2B
& action-only
& --
& 99.0 & 99.2 & 96.6 & 94.8 & 97.4 \\
VLANeXt-WAM
& WAN2.2-5B
& action-only
& --
& 95.0 & 99.6 & 96.4 & 91.2 & 95.6 \\
\midrule
\rowcolor[HTML]{EFEFEF}
\multicolumn{9}{l}{\textit{World Action Modeling}} \\
VLANeXt-WAM
& WAN2.2-5B
& action \& video
& fast connection
& 97.8 & 99.4 & 99.6 & 96.0 & 98.2 \\
VLANeXt-WAM
& WAN2.2-5B
& action \& video
& joint connection
& 95.2 & 99.8 & 97.0 & 98.2 & 97.6 \\
\bottomrule
\end{tabular}
\label{tab:wam_comparison}
\vspace{-2mm}
\end{table*}

\begin{table}[t]
\centering
\small
\caption{
Ablation of different backbone choices for VLANeXt on the LIBERO benchmark.
}
\begin{tabular}{l|ccccc}
\toprule
\textbf{Backbone}
& \multicolumn{5}{c}{\textbf{LIBERO}} \\
\cmidrule(lr){2-6}
& Spatial & Object & Goal & Long & Avg \\
\midrule
Qwen3.5-0.8B
& 95.4 & 99.4 & 95.4 & 96.4 & 96.7 \\
Qwen3.5-2B
& 97.8 & 100.0 & 98.0 & 97.4 & 98.3 \\
Qwen3.5-4B
& 97.4 & 99.6 & 97.2 & 95.2 & 97.4 \\
\midrule
Qwen3-VL-2B
& 99.0 & 99.2 & 96.6 & 94.8 & 97.4 \\
Qwen3-VL-4B
& 99.2 & 100.0 & 98.0 & 96.4 & 98.4 \\
Qwen3-VL-8B
& 98.0 & 99.4 & 97.4 & 95.8 & 97.7 \\
\bottomrule
\end{tabular}
\label{tab:backbone_ablation}
\vspace{-6mm}
\end{table}

\section{Beyond the Core Recipe: Extending VLANeXt to Emerging Paradigms}
\label{sec:new_paradigms}

\begin{figure}[t]
  \begin{center}
    \centerline{\includegraphics[width=\columnwidth]{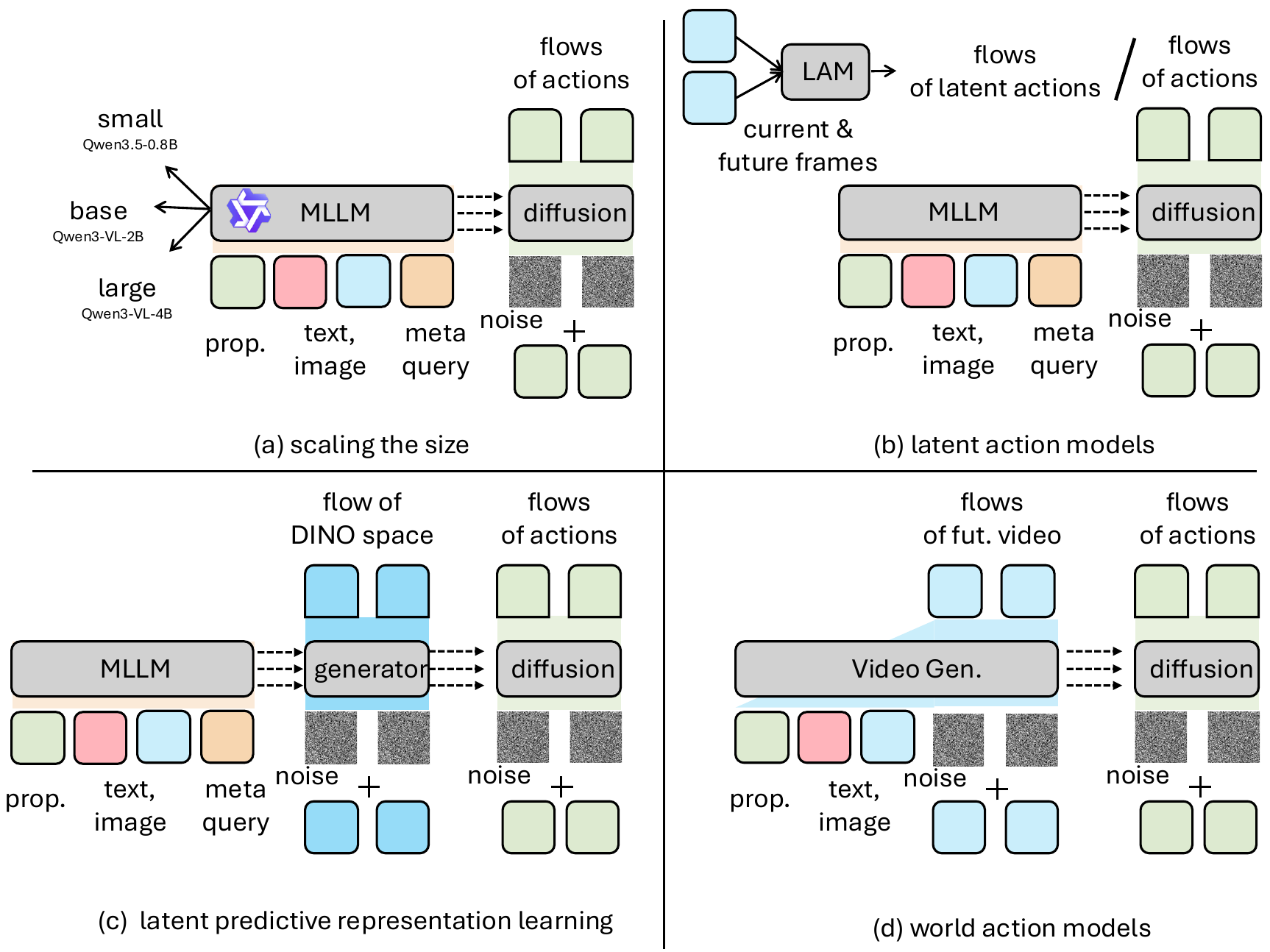}}
    \vspace{-1mm}
    \caption{
        \textbf{VLANeXt Family: Extending VLANeXt to Emerging Paradigms}. We extend the core VLANeXt recipe along four emerging directions: (a) model scaling with different MLLM backbone sizes; (b) latent-action pretraining with latent actions from current and future frames; (c) latent predictive representation learning through JEPA-style prediction in DINO space; and (d) world action modeling by jointly modeling future video and actions. These variants form the VLANeXt family, enabling us to examine the generality of the distilled design recipe across model scales and emerging paradigms.
    }
    \label{fig:emerging_paradigms}
  \end{center}
  \vspace{-8mm}
\end{figure}

The VLA field is rapidly evolving beyond the standard formulation, with several emerging paradigms introducing different priors for action learning. We focus on four representative directions: \textit{model scaling}, which introduces model-capacity priors~\cite{2025_tinyvla, 2026_vlm4vla}; \textit{latent-action pretraining}, which leverages action-like priors learned from unlabeled visual transitions~\cite{2024_LAPA, 2025_motus, 2026_latent-in-the-wild}; \textit{latent predictive representation learning}, which provides semantic priors through JEPA-style future representation prediction~\cite{2026_jepavla, 2026_leworldmodel}; and \textit{world action modeling}, which incorporates world-dynamics priors by jointly modeling future visual dynamics and robot actions~\cite{2026_dreamzero, 2026_fastwam, 2026_lingbot-va, 2026_kairos}. To examine whether the VLANeXt core recipe generalizes to these emerging paradigms, we instantiate them within the same framework, yielding \textbf{VLANeXt-S} and \textbf{VLANeXt-L} for model scaling, \textbf{VLANeXt-LAM} for latent-action pretraining, \textbf{VLANeXt-JEPA} for latent predictive representation learning, and \textbf{VLANeXt-WAM} for world action modeling. Together with the original VLANeXt, these variants form the \textbf{VLANeXt family}, as illustrated in Fig.~\ref{fig:emerging_paradigms}. As shown in Tab.~\ref{tab:core_recipe_generalization}, the \textit{core recipe remains effective across these variants despite their distinct model scales and sources of prior}, demonstrating that the identified design principles generalize beyond the standard VLA formulation.

\subsection{Model Scaling}
First, we study \textit{model scaling} in VLANeXt. Since the majority of model capacity in a VLA typically comes from its pretrained backbone, scaling VLANeXt can be naturally realized by varying the backbone size while keeping the remaining architecture and training recipe unchanged. Specifically, we replace the original Qwen3-VL-2B backbone~\cite{2025_qwen3-vl} with Qwen3.5-0.8B~\cite{2026_qwen3.5} and Qwen3-VL-4B~\cite{2025_qwen3-vl}, yielding \textbf{VLANeXt-S} (\textit{S} for small) and \textbf{VLANeXt-L} (\textit{L} for large), respectively. Together with the original VLANeXt model, these variants form a controlled scaling spectrum across different model capacities. By consistently adopting backbones from the Qwen family, we minimize confounding factors introduced by changes in backbone design, enabling a cleaner evaluation of whether the core recipe remains effective across model scales. An overview of this extension is shown in Fig.~\ref{fig:emerging_paradigms}(a), and more analysis can be seen in Sec.~\ref{sec:new_paradigms_evaluation}.

\subsection{Latent Action Modeling}
Second, we study \textit{latent-action pretraining}, which leverages action-like representations learned from visual transitions as a prior for downstream action learning~\cite{2024_LAPA, 2025_motus, 2026_latent-in-the-wild}. A latent action model (LAM) typically learns a compact representation of the transition between two observations. Specifically, a latent action encoder takes the current and future frames as input and extracts a latent action that captures the transition between them, resembling an inverse-dynamics mechanism. The resulting latent action is then combined with the current frame to reconstruct or predict the future frame. Since the future frame is available to the encoder while also serving as the prediction target, the latent space must be constrained to prevent the latent action from trivially encoding the target observation. Representative constraints include VAE~\cite{2013_vae}, which regularizes the latent space toward a Gaussian distribution; VQ-VAE~\cite{2017_vq-vae}, which discretizes the latent representation through vector quantization; and VICReg~\cite{2022_vicreg}, which encourages informative, diverse, and decorrelated latent representations, as explored in Latent-in-the-Wild~\cite{2026_latent-in-the-wild}.

We incorporate this paradigm into VLANeXt through a two-stage latent-action pretraining and real-action fine-tuning pipeline. We first train a LAM on the full LIBERO dataset~\cite{2023_libero} and use it to extract latent actions for all LIBERO trajectories. We then pretrain VLANeXt on the full dataset using these latent actions as supervision, before fine-tuning the pretrained model on each LIBERO suite with the corresponding ground-truth robot actions. This setup allows us to directly examine whether action-like priors learned from visual transitions can be transferred to downstream action learning with real action labels. An overview of this extension is shown in Fig.~\ref{fig:emerging_paradigms}(b), with further analysis provided in Sec.~\ref{sec:new_paradigms_evaluation}.

\subsection{Latent Predictive Representation Learning}

Next, we study \textit{latent predictive representation learning}, also referred to as JEPA-style predictive representation learning. This paradigm leverages semantic representations learned by pretrained representation models, such as DINO~\cite{2025_dinov3} and JEPA~\cite{2023_jepa}, as a prior for action learning. Rather than predicting future observations directly in pixel or token space, the model learns to predict their latent semantic representations and jointly optimizes this predictive objective with action learning. The underlying motivation is that effective robot control depends primarily on capturing task-relevant semantics, such as what changes should occur and where the interaction should take place, rather than reconstructing all low-level visual details.

We incorporate this paradigm into VLANeXt by introducing an additional predictor between the VLM backbone and the policy module, which predicts the latent representation of the future observation. This design naturally extends the world-modeling objective explored earlier in our core-recipe study, where an auxiliary predictor was used to predict Emu3.5~\cite{2025_emu3.5} tokens of the future image. Here, we retain the same general predictive formulation but replace the Emu3.5 tokenizer with the semantic encoder DINOv3~\cite{2025_dinov3}, shifting the prediction target from generative visual tokens to semantic latent representations. This controlled modification allows us to examine whether semantic predictive priors provide a more effective auxiliary signal for action learning. An overview of this extension is shown in Fig.~\ref{fig:emerging_paradigms}(c), with further analysis provided in Sec.~\ref{sec:new_paradigms_evaluation}.

\subsection{World Action Modeling}

Finally, we study \textit{world action modeling}~\cite{2026_dreamzero, 2026_fastwam, 2026_lingbot-va, 2026_kairos}. World modeling has long been explored as a way to provide predictive priors for robot learning, including the future-image token prediction objective investigated in our core-recipe study and the latent predictive representation learning paradigm discussed above. Here, we focus on a recently emerging formulation, World Action Models (WAMs), which replace the conventional VLM backbone with a pretrained video-generation model and jointly model future visual dynamics and robot actions. Since the video-generation backbone is pretrained on large-scale video data, it provides a strong world-dynamics prior that can be transferred to downstream action learning.

We incorporate this paradigm into VLANeXt by replacing the original VLM backbone with a video-generation backbone and jointly optimizing video generation and action prediction within the same framework. This extension allows us to examine whether the core VLANeXt recipe remains effective when action learning is built upon an explicit world-dynamics prior rather than a conventional VLM representation. An overview of this extension is shown in Fig.~\ref{fig:emerging_paradigms}(d), with further analysis provided in Sec.~\ref{sec:new_paradigms_evaluation}.

\section{Evaluating the Core Recipe Across Emerging Paradigms}
\label{sec:new_paradigms_evaluation}

\subsection{Settings}
In this section, we conduct further analysis and ablation studies for the four emerging paradigms introduced in Sec.~\ref{sec:new_paradigms}.
We follow the training and evaluation protocol described in Sec.~\ref{sec:core_recipe_evaluation}. Specifically, models are trained on LIBERO~\cite{2023_libero} and evaluated on all four suites: Spatial, Object, Goal, and Long. We report the success rate for each suite and their average.

\subsection{Ablation Analysis}

\textbf{Model Scaling.}
We first investigate model scaling by evaluating additional backbone sizes from both Qwen3.5~\cite{2026_qwen3.5} and Qwen3-VL~\cite{2025_qwen3-vl}. As reported in Tab.~\ref{tab:backbone_ablation}, increasing backbone capacity initially leads to clear improvements. Within the Qwen3.5 family, scaling from 0.8B to 2B parameters improves the average success rate from 96.7\% to 98.3\%, while within Qwen3-VL, scaling from 2B to 4B improves performance from 97.4\% to 98.4\%. However, the gains are not monotonic at larger scales: Qwen3.5-4B achieves 97.4\% and Qwen3-VL-8B achieves 97.7\%, both below their intermediate-scale counterparts. We hypothesize that this saturation is partly due to the limited scale and diversity of the LIBERO dataset, which may be insufficient to fully exploit the additional capacity of larger pretrained backbones.

\textbf{Latent-Action Constraints.}
We then investigate how different constraints on the latent action space affect latent-action pretraining. Following Latent-in-the-Wild~\cite{2026_latent-in-the-wild}, we compare three representative formulations: VAE~\cite{2013_vae}, VQ-VAE~\cite{2017_vq-vae}, and VICReg~\cite{2022_vicreg}. We additionally include a VLANeXt baseline trained for 30k steps, matching the total optimization duration of latent-action pretraining followed by action fine-tuning, to distinguish the benefit of latent-action priors from that of simply training for longer. As shown in Tab.~\ref{tab:lam_comparison}, increasing the baseline training duration from 10k to 30k steps does not improve performance, with the average success rate decreasing slightly from 97.4\% to 97.0\%. In contrast, the VQ-VAE variant achieves the best performance among the three latent-action formulations, reaching an average success rate of 97.6\% and improving particularly on the Long suite. These results suggest that the gain from VLANeXt-LAM is not explained by additional optimization alone, but depends on the action-like prior acquired during latent-action pretraining.
We hypothesize that the discrete bottleneck introduced by VQ-VAE encourages a compact representation of visual transitions and suppresses shortcut encoding, making the resulting latent actions more transferable to downstream action learning.

\textbf{Latent Predictive Objectives.}
We next examine the design of the predictive objective in VLANeXt-JEPA along two dimensions: the representation space used as the prediction target and the temporal horizon of prediction. For the representation space, we compare semantic features extracted by DINOv3~\cite{2025_dinov3} against the Emu3.5~\cite{2025_emu3.5} visual tokens considered in our core-recipe study. For the temporal target, we compare predicting the final observation within the current action horizon against predicting the final observation of the entire episode. As shown in Tab.~\ref{tab:jepa_comparison}, predicting the horizon-final representation achieves an average success rate of 97.7\%, outperforming both the episode-final variant (96.6\%) and the action-only VLANeXt baseline (97.4\%). This result indicates that predictive representations are most useful when their temporal scope remains closely aligned with the action horizon, whereas predicting a substantially more distant future state can introduce an overly difficult or weakly aligned auxiliary objective.
The comparison with the Emu3.5-token objective further shows that prediction in the DINOv3 semantic space is more effective than prediction in the generative visual-token space, suggesting that task-relevant semantics provide a more useful predictive prior for action learning than low-level visual reconstruction targets.

\textbf{World Action Modeling.}
Finally, we study how video generation should interact with action prediction in VLANeXt-WAM. We compare three variants based on the same WAN2.2-5B~\cite{2025_wan} backbone: an action-only model without video-generation supervision and two models using the fast and joint connections introduced in Fast-WAM~\cite{2026_fastwam}. In the fast formulation, action prediction does not directly attend to future video tokens, whereas the joint formulation allows interactions between the action and future-video streams. As shown in Tab.~\ref{tab:wam_comparison}, using the video backbone for action prediction alone achieves an average success rate of only 95.6\%. Introducing video-generation and action-learning objectives together substantially improves performance, reaching an average success rate of 98.2\% with the fast connection and 97.6\% with the joint connection. Importantly, the fast formulation performs best despite preventing action prediction from directly attending to future video tokens. This suggests that the main benefit of world action modeling comes from transferring the world-dynamics prior learned through the video-generation objective, rather than requiring action prediction to explicitly consume the large-scale generated future representations.

\section{Conclusion}
In summary, this work takes a step toward a more systematic understanding of VLA models. Rather than introducing another standalone architecture, we reexamine the VLA design space under a unified framework and evaluation protocol, spanning foundational components, perception essentials, and action modeling perspectives. Our study shows that strong VLA performance can emerge from principled design choices throughout the pipeline, including how the VLM interacts with the policy module, how perceptual signals such as multi-view observations and proprioception are integrated, and how temporal structure in actions is modeled.

Beyond the standard VLA formulation, we further examine whether the distilled core recipe generalizes to emerging paradigms, including model scaling, latent-action pretraining, latent predictive representation learning, and world action modeling. The resulting VLANeXt family demonstrates that these design principles remain effective across different model scales and sources of prior, suggesting that the core recipe provides a broadly applicable foundation as VLA paradigms continue to evolve.

By releasing a unified and lightweight framework, we aim to facilitate reproducible studies and systematic exploration of future VLA designs. Extending this perspective to more diverse embodiments, large-scale mid-training, failure recovery, and richer forms of prior knowledge remains an important direction for future research.

\bibliographystyle{IEEEtran}
\bibliography{VLANeXt}

\end{document}